\documentclass[journal]{IEEEtran}
\usepackage{amssymb}
\usepackage{graphicx}
\usepackage{amsmath}
\usepackage{amsthm}
\usepackage{amsfonts}
\usepackage{mathrsfs}
\usepackage[]{caption2}
\usepackage{bm}
\usepackage{enumerate}
\usepackage{booktabs}
\usepackage[]{graphicx}
\usepackage{stfloats}
\usepackage{supertabular}
\usepackage{multirow}
\usepackage{color}
\allowdisplaybreaks

\usepackage{booktabs}
\usepackage{threeparttable}

\usepackage{graphicx}
\usepackage{subfigure}
\usepackage{algorithm}
\usepackage{algpseudocode}
\usepackage{amsmath}

\ifCLASSINFOpdf
\else
\fi

\begin{document}

\title{Properties and Potential Applications of Random Functional-Linked Types of Neural Networks}

\author{
        Guang-Yong Chen,
        Yong-Hang Yu,\\
        Min Gan,~\IEEEmembership{~Senior Member,~IEEE},
        C. L. Philip Chen,~\IEEEmembership{~Fellow,~IEEE},
        Wenzhong Guo\\
\thanks{

G.-Y. Chen, M. Gan, Y.-H. Yu and W.-Z. Guo are with the College of Computer and Data Science, Fuzhou University, Fuzhou 350116, China (e-mail: cgykeda@mail.ustc.edu.cn; aganmin@gmail.com).

C. L. Philip Chen is with the School of Computer Science and Engineering, South China University of Technology, Guangzhou 510641, China (e-mail: philip.chen@ieee.org)

}}

\markboth{}%
{Shell \MakeLowercase{\textit{et al.}}: Frequency Principle in Broad Learning System}

\maketitle

\begin{abstract}
Random functional-linked types of neural networks (RFLNNs), e.g., the extreme learning machine (ELM) and broad learning system (BLS), which avoid suffering from a time-consuming training process, offer an alternative way of learning in deep structure. The RFLNNs have achieved excellent performance in various classification and regression tasks, however, the properties and explanations of these networks are ignored in previous research. This paper gives some insights into the properties of  RFLNNs from the viewpoints of  frequency domain, and discovers the presence of frequency principle in these networks, that is, they preferentially capture low-frequencies quickly and then fit the high frequency components during the training process. These findings are valuable for understanding the RFLNNs and expanding their applications. Guided by the frequency principle, we propose a method to generate a BLS network with better performance, and design an efficient algorithm for solving Poison's equation in view of the different frequency principle presenting in  the Jacobi iterative method and BLS network.
\end{abstract}

\begin{IEEEkeywords}
Random functional-linked types of neural networks (RFLNNs), Broad learning system (BLS), Frequency principle, Fourier domain, Jacobi iterative method.
\end{IEEEkeywords}

\IEEEpeerreviewmaketitle

\section{Introduction}

\subsection{Background and related work}
\IEEEPARstart{D}{eep} structure neural networks have achieved breakthrough success in a wide range of tasks such as image classification \cite{zhu2020deep}, segmentation \cite{badrinarayanan2017segnet}, and speech recognition \cite{hinton2012deep}. However, deep neural networks (DNNs) usually suffer from a time-consuming training process, involving tuning a huge number of hyperparameters and complicated structure.  Moreover, a complete retraining process is required when the structure is not sufficient to model the dataset. RFLNNs \cite{igelnik1995stochastic,chen2017broad,huang2006extreme} provide an alternative scheme that removes the drawbacks of a long learning and retraining process, and have achieved excellent performance in various application fields \cite{gong2021research,jin2021pattern,chu2019weighted}. At early stage, RFLNNs were  developed from flat networks. Chen \emph{et al.} \cite{chen1999rapid} proposed a random vector functional linked neural network (RVFLNN) based on the study of single layer feedforward neural networks (SLFNNs), and designed a fast learning algorithm to find the optimal weights of the networks. RVFLNN not only avoids a time-consuming training process but also achieves a universal approximation performance in function approximation \cite{pao1994learning,igelnik1995stochastic}. To adapt to the explosive growth of data in size and the sharp increase in dimension, Chen \& Liu \cite{chen2017broad,chen2018universal} proposed a broad learning system (BLS) based on the basic idea of RVFLNN, and developed an incremental learning algorithm for fast remodeling in broad expansion without a retraining process. Huang \emph{et al.} \cite{huang2006extreme} chose infinitely differentiable functions as the activation functions in the hidden layer of the SLFNNs, and proposed the extreme learning machine (ELM), which only tunes the parameters of output layer (linking the hidden layer to the output layer). The RVFLNN, ELM and BLS are different types of RFLNNs, which have been widely used for various tasks such as data analysis \cite{issa2021emotion,yang2021incremental}, traffic predictor \cite{liu2020training,xu2020recurrent}, time series analysis \cite{han2019structured}, image detection and processing \cite{ye2020adaptive,wang2021hybrid,chu2020hyperspectral}.

Recently, various improvements of the BLS and ELM have been presented to adapt to complex learning tasks. Feng \& Chen \cite{feng2020fuzzy} merged the Takagi-Sugeno fuzzy system into BLS, which achieved state-of-the-art performance compared to the nonfuzzy and neuro-fuzzy methods. Jin \emph{et al.} \cite{jin2018regularized,jin2018discriminative}, Bal \emph{et al.} \cite{bal2020wr}, and Gan \emph{et al.} \cite{gan2020weighted} introduced different regularizers to obtain robust BLS networks and robust ELM networks for different learning tasks.
Due to the powerful capacity of the deep structure, various variants of RFLNN combined with deep structure are proposed in different application fields. For example, in \cite{chen2018universal,zhang2022analysis,chu2022broad}, the BLS variants that use the recursive feature nodes and the cascade of feature mapping, resulting in the recurrent-BLS, gated-BLS, and CFEBLS. Yao \emph{et al.} \cite{yao2017deep,ma2014study} proposed a deep structure of ELM, which consists of several different level networks, to improve the effectiveness to deal with noisy data. Liu \emph{et al.} \cite{liu2020stacked} developed a stacked BLS by adding the ``neurons'' and ``layers'' dynamically during the training process for multilayer neural networks.

These RFLNNs (flat or deep) have achieved excellent performance in nature language processing, image classification, time-series forcasting and etc., however, the underlying principle of these random networks and why they can generalize well are still unclear. Exploring these problems is of great significance for understanding different types of RFLNNs, and has important enlightenment for expanding them to a wider range of applications.

\subsection{Motivation and Contribution}
Similar to DNNs, the RLFNNs are regarded as a black-box inference. Although researchers have made different structural modifications to random networks to adapt different applications, the lack of theoretical analysis remains an important factor limiting their development. This motivates us to give some insights into the widely used RFLNNs from frequency domain, including the ELM, BLS and stacked BLS with deep structure.

Xu \emph{et al.} \cite{xu2019training,xu2019frequency} found that the training process of DNNs initialized with small values of parameters usually fits the target functions from low-frequencies to high-frequencies, named frequency principle. Based on this valuable principle, researchers gave some explanations for the phenomenon of early stop and generalization puzzle in deep networks, and designed various efficient techniques for scientific computing problems \cite{xu2018frequency}. Motivated by this principle in deep networks, in this paper we explore the properties of the RFLNNs from the perspective of the frequency domain. This is valuable for readers to understand how different RFLNNs (e.g., the BLS and stacked BLS) work, and their differences, and their shortcomings compared to the DNNs. Moreover, the underlying principle is of great inspiration for extending RFLNNs to a wider range of applications.

The main contributions of this paper are listed as follows:

1) Explore the properties of the ELM, BLS and stacked BLS from the perspective of frequency domain, and find that the frequency principle holds in these random neural networks, i.e., they preferentially capture the low-frequencies and then gradually fit the high-frequency components. In addition, we find that these random neural networks are prone to instability in fitting the high-frequencies compared to the DNNs; The fitting accuracy of  the stacked BLS with deep structure is improved when adding the first two BLS blocks, but the fitting results of each frequency are not improved by adding other BLS blocks, which may be a key problem of the deep BLS networks to be solved in future research.

2) By the frequency principle in RFLNNs, we design a method to generate the BLS network with better prediction performance using the fact that the BLS gradually captures the high-frequency components during the training process.

3) According to the different frequency principles presenting in Jacobi iterative method and random neural networks, we propose a more efficient algorithm (denoted as BLS-Jacobi) for solving Poisson's equation.

The rest of this paper proceeds as follows. In Section II, different RFLNNs, including the ELM, BLS and stacked BLS are introduced, and we explore the properties of these random neural networks from the perspective of frequency domain in Section III. According to the frequency principle presenting in RFLNNs, we propose a method to generate a BLS network with better performance in Section IV. In Section V, an efficient algorithm combining the advantages of BLS and the iterative Jacobi method is proposed for solving the Poison equations. Finally, the main conclusion and further discussions of the RFLNNs are presented.

\section{Random functional-linked neural networks}
In this section, we briefly introduce some random neural networks, and the corresponding learning algorithms.
\subsection{Random vector functional-linked neural networks}
The RVFLNN \cite{pao1994learning,igelnik1995stochastic} is first proposed to overcome the drawback of trapping in a local minimum and long time training of single layer feedforward neural networks, and achieves a universal approximation for continuous functions with fast learning property \cite{tyukin2009feasibility}. Chen \emph{et al.} \cite{chen1999rapid} formulated the random flat networks as linear systems and proposed a step-wise updating algorithm to remodel the high-volume and time-variety data in modern large data era.

Fig. 1 shows the basic characteristics of the RVFLNN. The weights $\mathbf W_h$ that link the input nodes and enhance nodes and the bias $\boldsymbol \beta_h$ are randomly generated and remain unchanged thereafter. The RVFLNNs can be formulated as:
\begin{equation}\label{2-1}
  \mathbf Y=\left[\mathbf X,\ \mathbf \xi(\mathbf X\mathbf W_h+\boldsymbol e_N\otimes\boldsymbol \beta_h)\right]\mathbf W,
\end{equation}
where $\mathbf\xi(\cdot)$ is a nonlinear activation function which acts on the element-wise, $\mathbf X\in R^{N\times D}$ is the input matrix, $\mathbf Y\in R^{N\times K}$ is the output matrix, $\boldsymbol e_N=(1,\cdots,1)^\text T$ is an $N$-dimensional vector, $\otimes$ represents the Kronecker product, and the weights $\mathbf W$ from the input nodes and enhance nodes to the output nodes are required to be optimized. Pao \emph{et al.} \cite{pao1994learning} utilized a conjugate gradient search algorithm to find the optimal weights. In \cite{chen1999rapid}, Chen \emph{et al.} proposed a fast stepwise updating algorithm, which can easily retrain the network when a new observation or a new neuron is added to the existing network.
\begin{figure}
  \centering
  \includegraphics[width=0.48\textwidth]{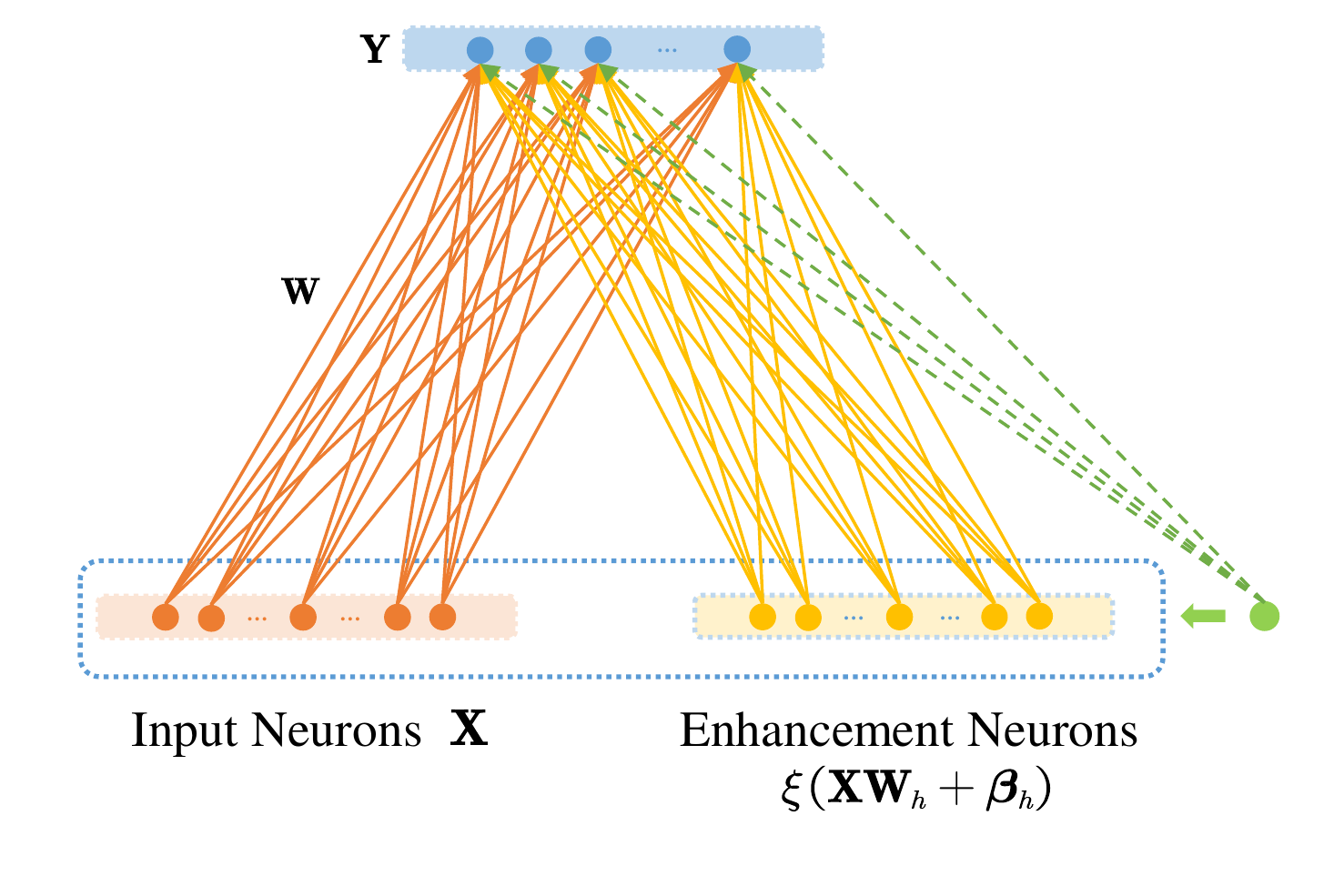}\\
  \caption{Basic structure of random vector functional-linked neural network}\label{Fig2-1}
\end{figure}
\subsection{Extreme learning machine}
The ELM is a class of random neural networks, which is first proposed by Huang \emph{et al.} \cite{huang2006extreme}. The hidden nodes of the ELM networks are randomly chose, and we just need to determine the output weights. An ELM networks with $L$ hidden nodes can be mathematically modelled as
\begin{equation}\label{2-2}
  \mathbf Y=\mathbf H \mathbf W
\end{equation}
where
\begin{equation*}
  \mathbf H=\left[
     \begin{array}{ccc}
       \xi(\boldsymbol w_1^\text T\mathbf x_1+b_1) & \cdots & \xi(\boldsymbol w_L^\text T\mathbf x_1+b_L) \\
       \vdots & \cdots & \vdots \\
       \xi(\boldsymbol w_1^\text T\mathbf x_N+b_1) & \cdots & \xi(\boldsymbol w_L^\text T\mathbf x_N+b_L) \\
     \end{array}
   \right]
\end{equation*}
is the hidden layer output matrix of the neural network. For given target observations, the output weights can be obtained by solving a least squares problem. Unlike the RVLNNs, there are not links from the input nodes to the output nodes in the ELM networks.
\subsection{Broad learning system}
\begin{figure}
  \centering
  \includegraphics[width=0.48\textwidth]{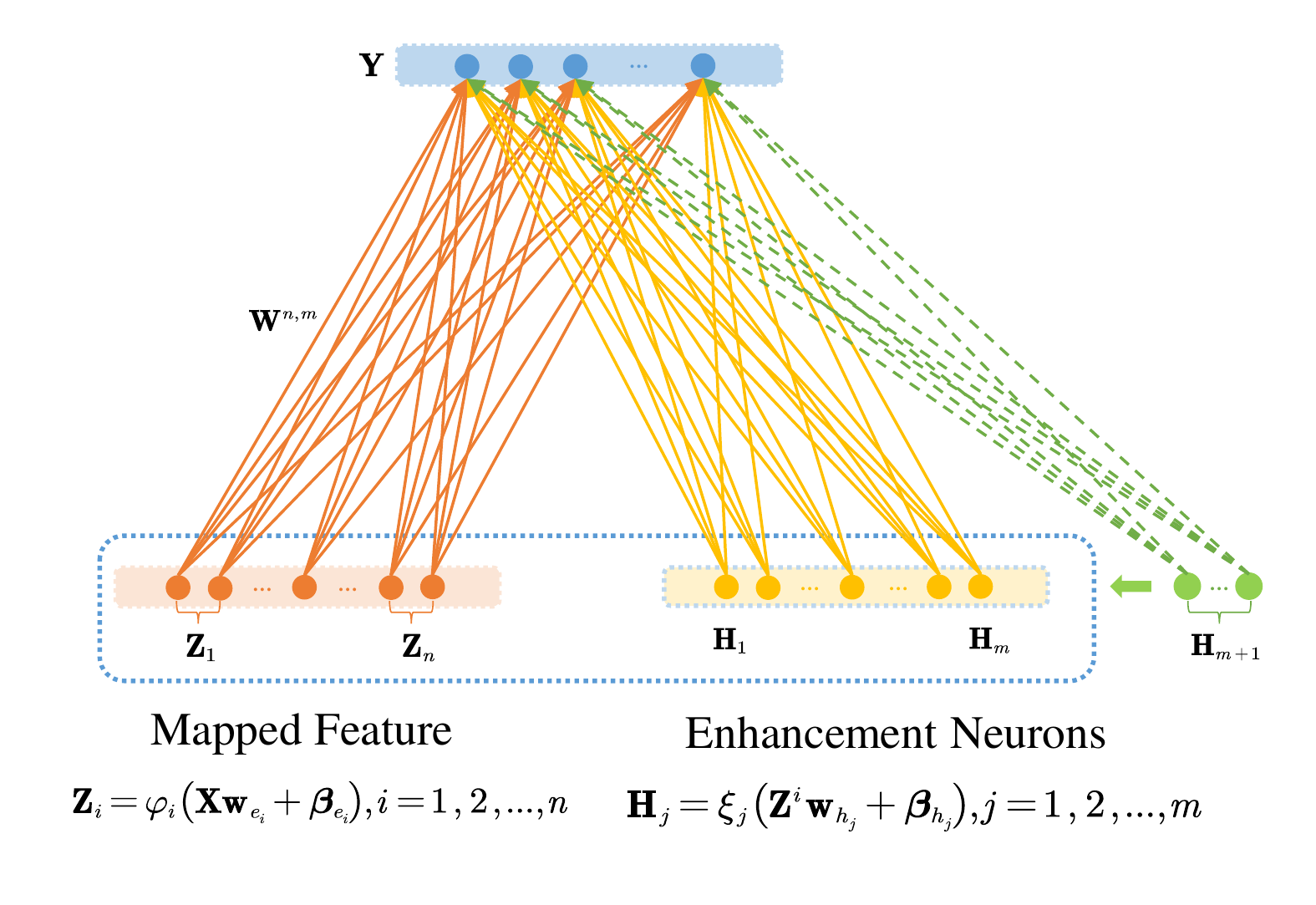}\\
  \caption{Basic structure of broad learning system}\label{Fig2-2}
\end{figure}
To adapt to large-scale data and high dimension encountered in complex learning tasks, Chen \emph{et al.} \cite{chen2017broad} proposed a BLS network based on the underlying idea of the RVFLNN. Unlike the RVFLNN, the BLS takes the features extracted form the raw data as input, which can effectively solve the problem of high dimensionality of the original data. In addition, it is convenient for the BLS to remodel in structure expansion without retraining.

Fig. 2 outlines the basic structure of the BLS. First, it maps the input to a set of features to form the feature nodes
\begin{equation*}
  \mathbf Z^i=\left[\mathbf Z_1,\cdots,\mathbf Z_i\right],
\end{equation*}
\begin{equation*}
  \mathbf Z_i=\boldsymbol \varphi_i(\mathbf X\mathbf w_{e_i}+\boldsymbol\beta_{e_i}).
\end{equation*}
The $j$th group of enhancement nodes can be constructed by the random mapping
\begin{equation*}
  \mathbf H_j=\boldsymbol \xi_j(\mathbf Z^i\mathbf w_{h_j}+\boldsymbol\beta_{h_j}).
\end{equation*}
Collecting the previous $j$ groups of the enhancement nodes to form $\mathbf H^j\equiv\left[\mathbf H_1,\cdots,\mathbf H_j\right]$. The parameters $\mathbf w_{e_i}$, $\boldsymbol \beta_{e_i}$, $\mathbf w_{h_j}$ and $\boldsymbol\beta_{h_j}$ are all randomly generated.

The parameters from feature nodes and enhancement nodes to the output nodes, denoted  as $\mathbf W^{n,m}$, can be obtained by solving the minimization problem
\begin{equation}\label{2-3}
  \min\limits_{\mathbf W^{n,m}}\left\|\mathbf A\mathbf W^{n,m}-\mathbf Y\right\|_v+\lambda\left\|\mathbf W^{n,m}\right\|_u,
\end{equation}
where $\mathbf A=\left[\mathbf Z^n, \mathbf H^m\right]$,  $u$ and $v$ represent some typical kinds of norm. When $u,v=2$, the optimal values can be obtained by the ridge regression method
\begin{equation}\label{2-4}
  \hat{\mathbf W}^{n,m}=(\lambda\mathbf I+\mathbf A^\text T\mathbf A)^{-1}\mathbf A^\text T\mathbf Y.
\end{equation}

Chen \emph{et al.} further develop an incremental learning algorithm for the new incoming input, the increment of the feature nodes and enhancement nodes. Here we just briefly introduce the incremental algorithm for the adding of enhancement nodes. For notational convenience, we assume that $\mathbf A^m=\left[\mathbf Z^n,\ \mathbf H^m\right]$ and $\mathbf A^{m+1}=\left[\mathbf A^m,\ \boldsymbol\xi(\mathbf Z^n\mathbf w_{h_{m+1}}+\boldsymbol\beta_{h_{m+1}})\right]$, then the output weights can be updated as follows after adding a group of enhancement nodes:
\begin{equation*}
  \mathbf W^{n,m+1}=\left[
                          \begin{array}{c}
                            \mathbf W^{n,m+1}-\mathbf D\mathbf B^\text T\mathbf Y \\
                            \mathbf B^\text T\mathbf Y \\
                          \end{array}
                        \right],
\end{equation*}
where \begin{small}$\mathbf D=(\mathbf A^m)^\dagger\boldsymbol\xi(\mathbf Z^n\mathbf w_{h_{m+1}}+\boldsymbol\beta_{h_{m+1}})$, $\mathbf C=\boldsymbol\xi(\mathbf Z^n\mathbf w_{h_{m+1}}+\boldsymbol\beta_{h_{m+1}})-\mathbf A^m\mathbf D$,\end{small} and
\begin{equation*}
  \mathbf B^\text T=\left\{\begin{array}{c}
                             \mathbf C^\dagger,\ \ \text{if}\  \mathbf C\neq\mathbf 0 \\
                             (\mathbf I+\mathbf D^\text T\mathbf D)^{-1}\mathbf B^\text T(\mathbf A^m)^\dagger,\ \  \text{otherwise}.
                           \end{array}
  \right.
\end{equation*}
\subsection{Stacked broad learning system}
In \cite{liu2020stacked}, Liu \emph{et al.} proposed a variant of BLS combined with deep structure network, named stacked BLS. The stacked BLS can be remodeled as an increment of ``neurons'' and ``layer'' dynamically during training process, which mainly includes two repeated steps (generating the BLS block and stacking it on the top of the stacked BLS). Fig. 3 shows the schematic plot of the stacked BLS. The corresponding incremental algorithm presented in \cite{liu2020stacked} calculates not only the linked parameters between the newly stacked block but also the linked parameters of the enhance nodes within the BLS block.
\begin{figure}[ht]
  \centering
  \includegraphics[width=0.48\textwidth]{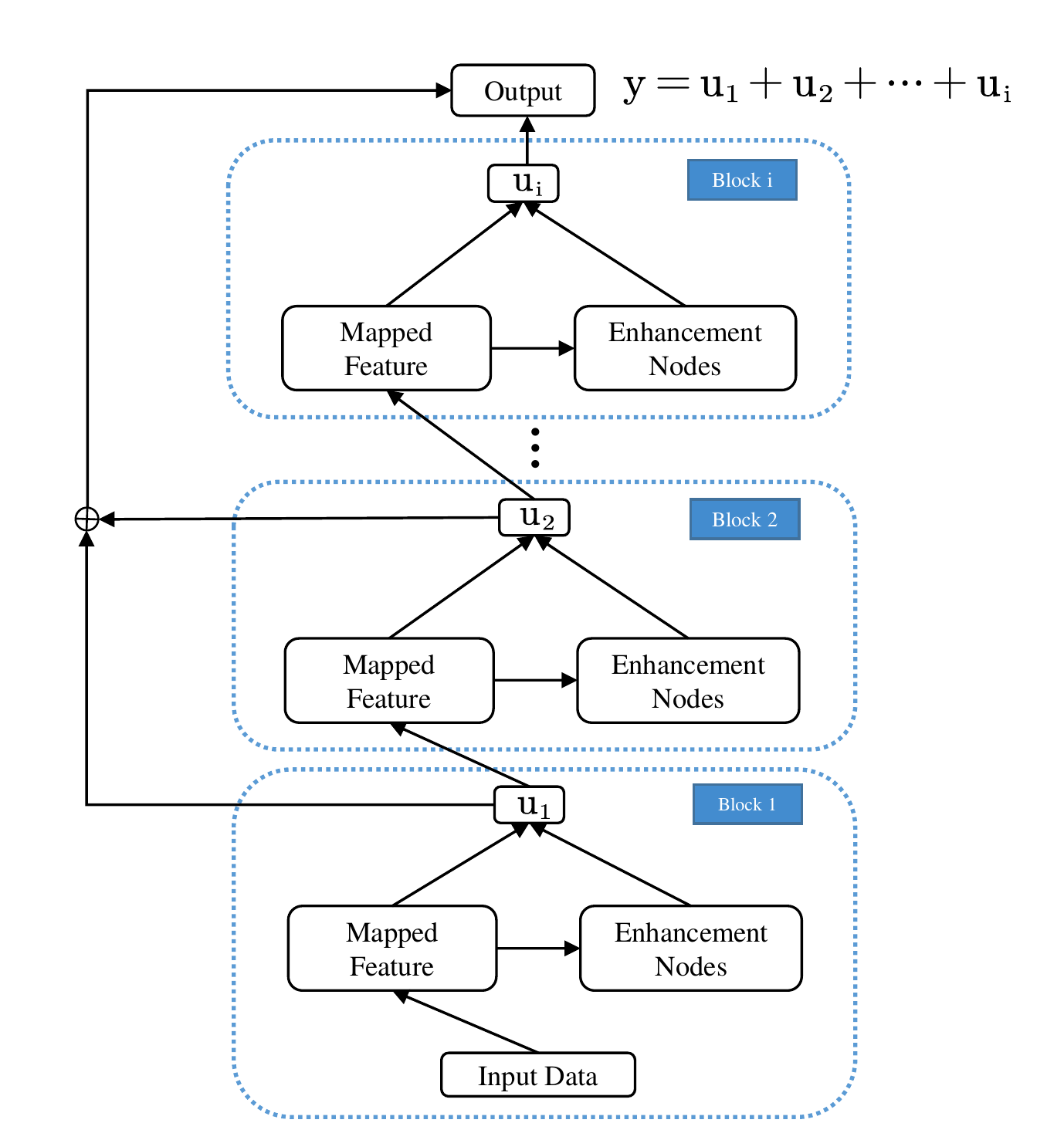}\\
  \caption{Illustration of the stacked broad learning system}\label{Fig2-3}
\end{figure}
\section{Properties of different types of RFLNNs}
In this section, we will explore the properties of RFLNNs (including ELM, BLS and stacked BLS) from the perspective of frequency domain. Frequency principle was proposed by Xu \emph{et al.} \cite{xu2019frequency} when they used the Fourier analysis tool to shed light on the DNNs. Some valuable tools were developed based on this principle for scientific research applications. Motivated by these, we provide some insights into the RFLNNs in the following part.
\subsection{Fitting sampling function}
\begin{figure}[h]
  \centering
  \includegraphics[width=0.48\textwidth]{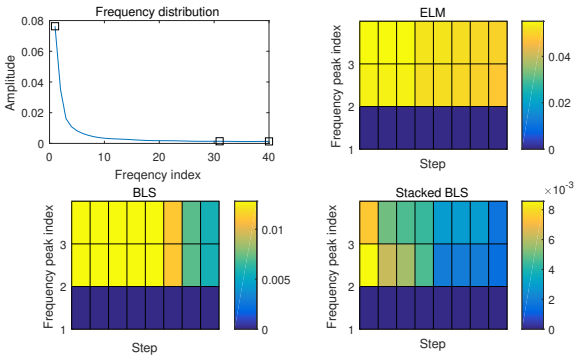}\\
  \caption{The upper left plot is the frequency distribution of the sampled data, including three peak points; The remaining plots are the fitting results observed at the three peak points during the training process of ELM, BLS and stacked BLS.}\label{Fig3-1}
\end{figure}

In this subsection, we use the ELM, BLS and stacked BLS networks to fit the sampling function
\begin{equation*}
  f(x)=\frac{\sin(x)}{x},
\end{equation*}
and then observe the fitting performance from frequency domain. The Fourier transformation of the sampling function is shown in the left top in Fig. \ref{Fig3-1}, where the physical frequency ($[0,20\pi]$) is replaced with the corresponding index (1:1:40). As discussed in \cite{xu2019training}, since frequency components except for the peaks are susceptible to the artificial periodic boundary condition implicitly applied in the Fourier transform \cite{percival1993spectral}, we only focus on the analysis of the convergence performance of the frequency peaks. The fitting results in the peak points obtained by different methods are shown in Fig. \ref{Fig3-1}. From Fig. \ref{Fig3-1}, we can observe that: the frequency principle holds in ELM, BLS and stacked BLS, i.e., they tend to fit low-frequency components preferentially and then gradually fit high-frequencies. In addition, the fitting results of the BLS and stacked BLS are better than ELM, which may attributed to that the BLS links both feature nodes and enhancement nodes to the output layer.

\emph{Remark}: In the simulation experiments, we found that for complex functions, this types of RFLNNs are difficult to capture the high-frequency information, which may be a major defect of this kind of random network.

\subsection{RFLNNs for image classification}
RFLNNs have achieved excellent performance in image classification task. Here, we will verify the frequency principle presenting in ELM, BLS and stacked BLS in the some popular datasets, including two sets of handwritten digital images and three persons pictures:
\begin{enumerate}
  \item The USPS dataset consists of 9298 handwritten digit images;
  \item The MNIST is a classical handwritten image dataset including 70000 digits;
  \item The EXYAB contains 2414 pictures taken by 38 persons with different expressions and illumination;
  \item The ORL dataset consists of 400 face pictures of 40 persons taken under different lights, times, and facial expressions;
  \item The UMIST dataset contains 575 pictures of 20 persons of different races, sexes, and appearances.
\end{enumerate}

For high-dimensional input, the curse of dimensionality is prone to occur when computing the Fourier transformation. As discussed in \cite{xu2018frequency,luo2021phase}, we just consider the first principal component of the input images when performing Fourier analysis.

Denote $\left\{\mathbf X\in \mathcal{R}^{N\times D}, \mathbf Y\in \mathcal{R}^{N\times K}\right\}$ as the input images, where $N$ is number of images, $D$ is the pixel number of a picture, and $K$ is the number of categories. We do the following preprocessing on $\mathbf X$ before performing Fourier analysis. First, the input data $\mathbf X$ is transformed to $\hat{\mathbf X}=\left[\hat{\mathbf x}_1^\text T,\cdots,\hat{\mathbf x}_N^\text T\right]$, where
\begin{equation*}
  \hat{\mathbf x}_j=\mathbf x_j-\frac{1}{N}\sum\limits_{k=1}^N\mathbf x_k,\ \ j=1,\cdots,N.
\end{equation*}
We then compute the principal component of the covariance matrix $\mathbf C=\hat{\mathbf X}\hat{\mathbf X}^\text T$, denoted as $\boldsymbol p$. Last, we project each observation  in the $\boldsymbol p$-direction and rescale the obtained element
\begin{equation*}
\bar{x}_{k}=\hat{x}_k \boldsymbol p,  k=1,2,\cdots,N,
\end{equation*}
\begin{equation*}
x_{k}^{\prime}=\frac{\bar{x}_{k} -\min_j \bar{x}_{j}}{\max_i(\bar{x}_{i}-\min_j \bar{x}_{j})}\in [0,1], \ i,j,k=1,\cdots,N.
\end{equation*}

Denote $\mathbf X^\prime=\left[x_1^\prime,\cdots,x_N^\prime\right]$ as the principal components. As discussed in \cite{chen2021frequency,xu2018frequency}, we consider the first dimension of the labels (denoted as $\boldsymbol y$) and do the nonuniform fast Fourier transform (NUFFT) on $\boldsymbol X$, which yields
\begin{equation*}
F[y](\alpha_{i})=\frac{1}{N}\sum_{k=0}^{N} y_{k} e^{-\rho i x_{k}^{\prime} \alpha_{i}},\ \  i= 1,2,\cdots,z,
\end{equation*}
where $\rho$ represents the frequency range and $\alpha_{i} \in \mathbb{Z}$ is the frequency index. After processing the above steps, the original training data of the networks can be formulated as follows:
\begin{equation*}\small
D_{y}=\{(\alpha_{1},F[y](\alpha_{1})), \cdots, (\alpha_{z},F[y](\alpha_{z}))\}.
\end{equation*}

A similar transformation is carried out on the output of the random neural networks (denoted by $\mathbf \Psi$), we have
\begin{equation*}\small
D_{t}=\{(\alpha_{1},F[t](\alpha_{1})), \cdots, (\alpha_{z},F[t](\alpha_{z}))\},
\end{equation*}
where $t$ is the first column of $\mathbf \Psi$. The relative error defined as follows is used to evaluate the fitting performance of different networks during the training process:
\begin{equation}\label{3-1}
  \Delta D(\alpha_i)=\frac{\left|\left|F[t^k](\alpha_i)\right|-\left|F[y](\alpha_i)\right|\right|}{\left|F[y](\alpha_i)\right|},
\end{equation}
where $t^k$ is obtained at the $k$th step.

\begin{figure}[h]
  \centering
  \includegraphics[width=0.48\textwidth]{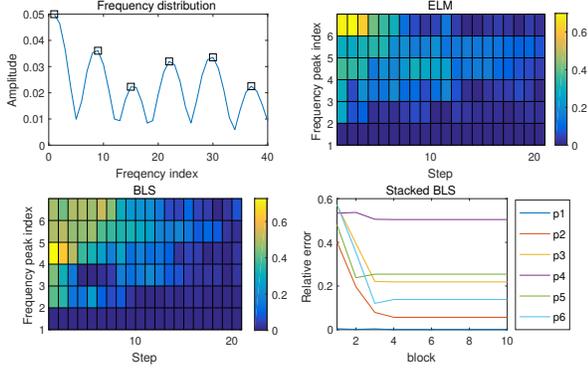}\\
  \caption{Fitting results of the UMIST data. The upper left plot is the frequency distribution of the sampled data. The remaining plots are the fitting results observed at each peak point of the distribution. The bottom right plot is the fitting error at each frequency point in the process of increasing the BLS blocks}\label{Fig3-2}
\end{figure}

\begin{figure}[h]
  \centering
  \includegraphics[width=0.48\textwidth]{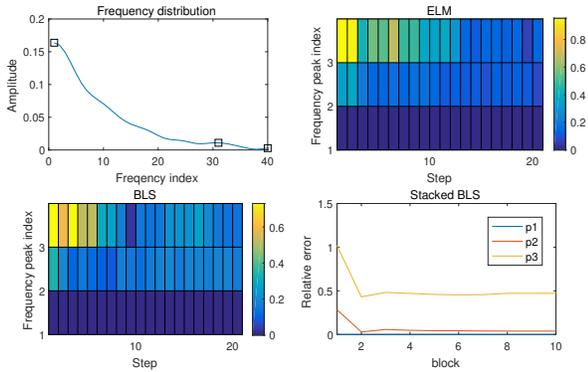}\\
  \caption{Fitting results of the USPS data.}\label{Fig3-3}
\end{figure}

\begin{figure}[h]
  \centering
  \includegraphics[width=0.48\textwidth]{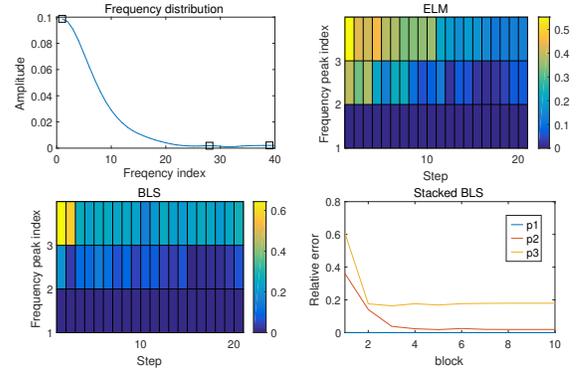}\\
  \caption{Fitting results of the MNIST data.}\label{Fig3-4}
\end{figure}

\begin{figure}[h]
  \centering
  \includegraphics[width=0.48\textwidth]{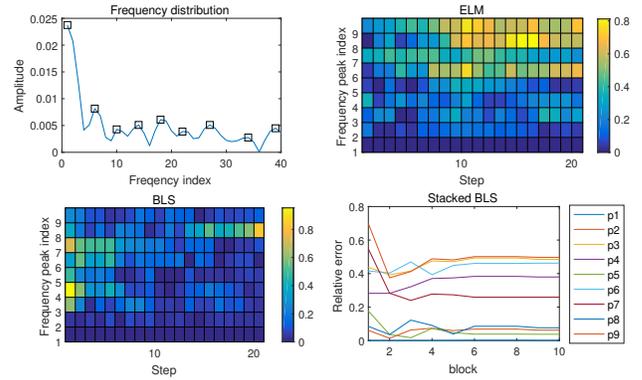}\\
  \caption{Fitting results of the EXYAB data.}\label{Fig3-5}
\end{figure}

\begin{figure}[h]
  \centering
  \includegraphics[width=0.48\textwidth]{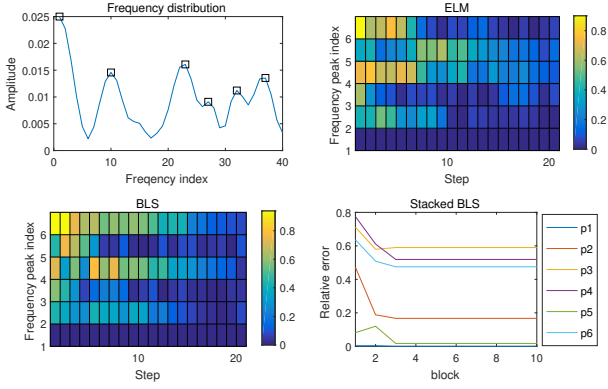}\\
  \caption{Fitting results of the ORL data.}\label{Fig3-6}
\end{figure}

Figs. \ref{Fig3-2}-\ref{Fig3-6} show the simulation results obtained by different random neural networks fitting different datasets. The frequency distributions shown in Figs. \ref{Fig3-2}-\ref{Fig3-6} suggest that the handwritten datasets USPS and MNIST are mainly dominated by low-frequencies, while the other three face datasets contains more high-frequency informations.

From the fitting results shown in Figs. \ref{Fig3-2}-\ref{Fig3-6}, we can observe that: these random types of neural networks usually capture the low-frequencies quickly and then gradually fit the high-frequency components.  The ELM and BLS networks are more prone to instability when fitting the high frequencies. The fitting results of the stacked BLS is improved when adding the first two BLS blocks, while continuing to deepen the network contributes little to the fitting accuracy. In order to show this phenomenon more intuitively, we use the relatively error curve of each peak frequency to illustrate it, as shown in Figs. \ref{Fig3-2}-\ref{Fig3-6}.

The simulation results confirm the existence of frequency principle presenting in the RFLNNs, which provides an important perspective to understand the properties of these random neural networks and  is of significance to expand their applications in more research fields.
\section{An improved method to generate BLS networks with better prediction performance}
The frequency principle presenting in RFLNNs is of great important for readers to understand the random networks and  to make some improvements. In \cite{chen2017broad,liu2020stacked}, the parameters of feature nodes and enhancement nodes in the BLS network are always generated in a fixed distribution interval during the training process, which is inconsistent with the frequency principle of the network. In this subsection, we will propose a method to generate a BLS network with better prediction performance according to the fact that the BLS gradually captures the high-frequency components during the training process.

In the following part, we first take the tanh function as an example to analyze the influence of parameters on the activation function from the perspective of the frequency domain. For convenience, one-dimensional case is discussed here. Performing Fourier analysis  on the function
\begin{equation*}
  \sigma(wx+b)=\tanh(wx+b)=\frac{e^{wx+b}-e^{-(wx+b)}}{e^{wx+b}-e^{-(wx+b)}},
\end{equation*}
we have
\begin{equation}\label{4-1}
  \hat{\sigma}(wx+b)(\zeta)=\frac{2\pi i}{|w|}\exp(\frac{ib\zeta}{w})\frac{1}{\exp(-\frac{\pi\zeta}{2w})-\exp(\frac{\pi\zeta}{2w})}
\end{equation}

Assume $\frac{\pi\zeta}{2w}>0$, here we focus on the high frequencies, i.e., $\zeta$ is large, which allows (\ref{4-1}) to be approximate by
\begin{equation}\label{4-2}
  \hat{\sigma}(wx+b)(\zeta)\approx \frac{2\pi i}{|w|}\exp(\frac{ib\zeta}{w})\exp(-\frac{\pi\zeta}{2w}).
\end{equation}
Similarly, when $\frac{\pi\zeta}{2w}<0$, we can obtain the approximated equation:
\begin{equation}\label{4-3}
  \hat{\sigma}(wx+b)(\zeta)\approx \frac{2\pi i}{|w|}\exp(\frac{ib\zeta}{w})\exp(\frac{\pi\zeta}{2w}).
\end{equation}

\begin{figure}
  \centering
  \includegraphics[width=0.48\textwidth]{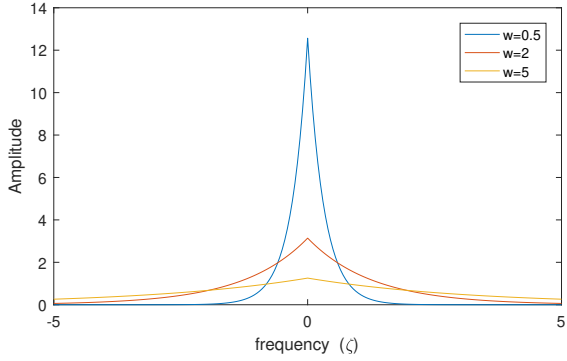}\\
  \caption{The spectrograms for different parameters w}\label{Fig4-4}
\end{figure}

Since BLS gradually captures high-frequency components, it is naturally hoped that the enhancement nodes and feature nodes added in the training process can provide more high-frequency information. According to Equations (\ref{4-2}) and (\ref{4-3}), a large $w$ is more beneficial to generate high frequency information, which can also be seen intuitively from Fig. \ref{Fig4-4}. Therefore, in this paper, we use a dynamically expanded interval to generate the parameters $w$ instead of using a fixed interval during the training process (e.g., the fixed interval [-1, 1] was used in \cite{chen2017broad,chen2018universal}).

\begin{figure}
  \centering
  \includegraphics[width=0.48\textwidth]{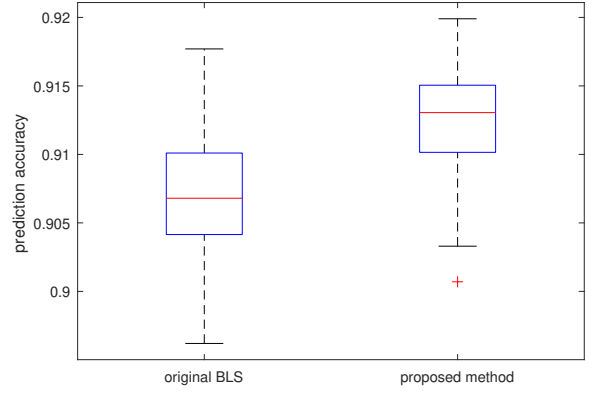}\\
  \caption{Comparisons of prediction accuracies of BLS networks generated by two different methods on MNIST dataset}\label{Fig4-1}
\end{figure}

\begin{figure}
  \centering
  \includegraphics[width=0.48\textwidth]{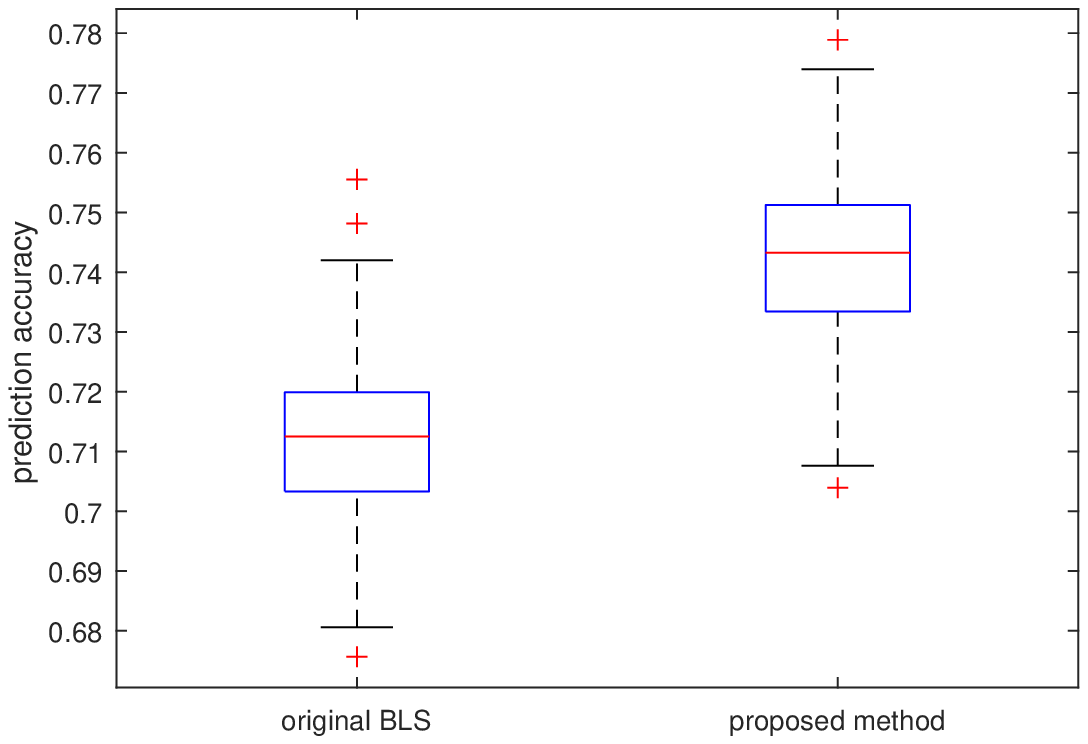}\\
  \caption{Comparisons of prediction accuracies of BLS networks generated by two different methods on EXYAB dataset}\label{Fig4-2}
\end{figure}

\begin{figure}
  \centering
  \includegraphics[width=0.48\textwidth]{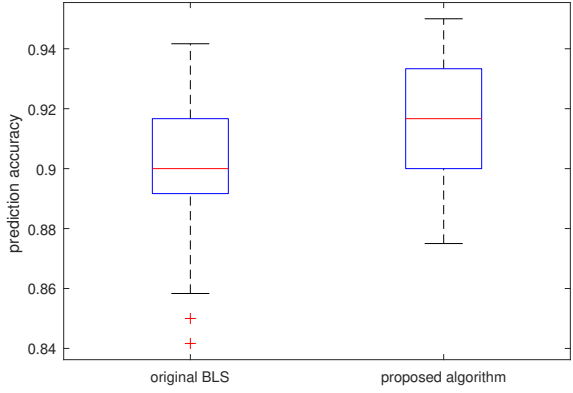}\\
  \caption{Comparisons of prediction accuracies of BLS networks generated by two different methods on ORL dataset}\label{Fig4-3}
\end{figure}

Next, we will verify the effectiveness of the proposed method to generate the BLS networks based on the frequency principle on different datasets (including the handwritten digital image dataset MNIST, face pictures dataset EXYAB and face pictures dataset ORL). In order to exclude the influence of random factors, we randomly run the original BLS training algorithm and the proposed method 100 times respectively. The boxplots of prediction accuracy obtained by the two methods are shown in Figs. \ref{Fig4-1}-\ref{Fig4-3}. From the Figures, it is easy to observe that the BLS network generated by the proposed method usually achieves higher prediction accuracy than the original method. The comparison results suggest that the proposed algorithm that takes into account the frequency principle presenting in the BLS can effectively improve the defect of insufficient fitting of high-frequency components, and achieves better performance than the original training algorithm.

\section{A novel method for solving Poisson's equation}
Poisson's equation is an important class of partial differential equations, which appears in a wide range of theoretical and application fields. Jacobi iterative algorithm is a conventional and effective method to solve these problems. In \cite{xu2018frequency,liu2020multi}, Xu \emph{et al.} designed a method (named DNN-Jacobi) for Poisson's equation according to the different frequency principle presented in DNN and convention method.
\begin{figure}
  \centering
  \includegraphics[width=0.48\textwidth]{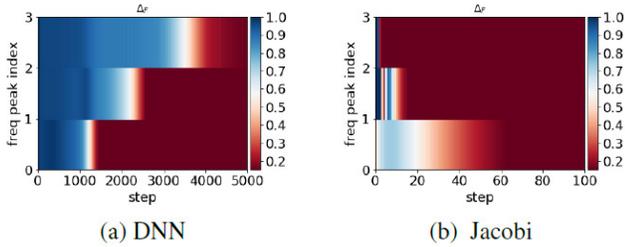}\\
  \caption{Frequency principle presented in DNN and Jacobi iterative method for solving Poisson's equation redrawn from \cite{xu2018frequency}}\label{Fig5-1}
\end{figure}

From Fig. \ref{Fig5-1}, we can observe that when using the Jacobi iterative method, the high-frequencies converge much faster than the low frequencies, which is opposite to the frequency principle presenting in DNN or BLS. Therefore, it is natural for us to combine the neural networks and Jacobi method to solve the Poisson's equation. Considering that DNN suffers from a time-consuming training process, in this section, we design a method, named BLS-Jacobi, based on the frequency principle presenting in BLS. The BLS-Jacobi method mainly consists of two parts: First, the BLS with $M$ incremental steps is used to solve the Poisson's equation; Then we use the output of the BLS as the initial value for the Jacobi iterative method.

Here we consider a 1-dimension Poisson's equation
\begin{eqnarray}\label{5-1}
&&  \Delta u(x)=g(x),\ x\in\Omega=(-1,1)\\
&&  u(x)=0,\ x=-1,1,\nonumber
\end{eqnarray}
and a 2-dimension Poisson's equation
\begin{equation}\label{5-2}
  \left\{\begin{array}{l}
           -\Delta u=f(x,y),\ (x,y)\in G=(0,1)\times(0,1) \\
           u|_{\partial G}=\left\{\begin{array}{l}
                                    0,\ x=0\ \text{ or}\  y=0 \\
                                    y^2,\ x=1 \\
                                    x^2,\ y=1
                                  \end{array}
           \right.
         \end{array}
  \right.,
\end{equation}
where $g(x)=\sin(x)+4\sin(4x)-8\sin(8x)+16\sin(24x)$, $f(x,y)=-2(x^2+y^2)$ and $\Delta$ is the Laplace operator.

As discussed in \cite{xu2018frequency}, the Poisson's equations here are solved by central differencing scheme. For example, Eq (\ref{5-1}) is discretized into the following form:
\begin{equation}\label{5-3}
  -\Delta u_i=-\frac{u_{i+1}-2u_{i}+u_{i-1}}{(\Delta x)^2}=g(x_i),\ i=1,2,\cdots,n.
\end{equation}
To express more compactly, Eq (\ref{5-3}) can be written in a matrix form:
\begin{equation}\label{5-4}
  \mathbf A \boldsymbol u=\boldsymbol g,
\end{equation}
where
\begin{equation*}
  \mathbf A=\left(
              \begin{array}{cccccc}
                2 & -1 & 0 & 0 & \cdots & 0 \\
                -1 & 2 & -1 & 0 & \cdots & 0 \\
                0 & -1 & 2 & -1 & \cdots & 0 \\
                \vdots & \vdots & \cdots &  &  & \vdots \\
                0 & 0 & \cdots & 0 & -1 & 2 \\
              \end{array}
            \right)_{n-1\times n-1}
\end{equation*}

\begin{equation*}
  \boldsymbol u=\left(
                  \begin{array}{c}
                    u_1 \\
                    u_2 \\
                    \vdots \\
                    u_{n-2} \\
                    u_{n-1} \\
                  \end{array}
                \right),\ \
                \boldsymbol g=(\Delta x)^2\left(
                                            \begin{array}{c}
                                              g_1 \\
                                              g_2 \\
                                              \vdots \\
                                              g_{n-2} \\
                                              g_{n-1} \\
                                            \end{array}
                                          \right).
\end{equation*}

\begin{table}[h]
  \centering
  \begin{center}
  \tabcolsep 40pt
  \caption{Time-consuming comparison of three different algorithms to achieve the preset accuracy (1-dimension Poisson's equation, Unit: s)}
  \end{center}
  \label{Tab5-1}
  \begin{tabular}{l c c c}
  \toprule
  Preset accuracy&Jacobi&DNN-Jacobi&BLS-Jacobi\\
  \midrule
  1e-1&0.0647&0.5733&0.0236 \\
  \\
  1e-2&0.5494&10.0101&0.0239\\
  \\
  1e-3&1.3003&10.3711&0.0481 \\
  \\
  1e-4&2.3566&11.6533&0.0774\\
  \\
  1e-5&3.1058&13.9477&0.2549 \\
  \\
  1e-6&4.8168&15.3823&0.8736\\
  \bottomrule
  \end{tabular}
\end{table}

\begin{table}[h]
  \centering
  \begin{center}
  \tabcolsep 40pt
  \caption{Time-consuming comparison of three different algorithms to achieve the preset accuracy (2-dimension Poisson's equation, Unit: s)}
  \end{center}
  \label{Tab5-2}
  \begin{tabular}{l c c c}
  \toprule
  Preset accuracy&Jacobi&DNN-Jacobi&BLS-Jacobi\\
  \midrule
  1e-1&0.4537&1.4914&0.0479 \\
  \\
  1e-2&6.1470&5.5628&0.0479\\
  \\
  1e-3&13.5442&9.4029&0.0479 \\
  \\
  1e-4&21.0617&12.9887&0.6758\\
  \\
  1e-5&30.0273&24.8210&2.0716 \\
  \\
  1e-6&39.1745&33.5102&8.2575\\
  \bottomrule
  \end{tabular}
\end{table}

The Jacobi iterative algorithm, DNN-Jacobi algorithm and the proposed BLS-Jacobi algorithm are adopted to solving the Poisson's equations (Eq (\ref{5-1}) and Eq (\ref{5-2})). Tables I and II list the time required to achieve a preset accuracy when using the three different algorithms to solve the 1-dimensional and 2-dimensional Poisson's equations. As shown in Tables I and II, we can observe that: To achieve the same accuracy, the Jacobi iterative method consumes significantly more time than the proposed BLS-Jacobi method; The DNN-Jacobi method is also more time-consuming than the BLS-Jacobi due to the complex training process of DNN, but the result is better than the pure Jacobi iterative method when solving the 2-dimensional Poisson's equation (for 2-dimensional case, the matrix size in discretized equation  is $(n-1)^2\times (n-1)^2$, which is much larger than that of 1-dimensional Poisson's equation); The BLS-Jacobi method takes advantage of the fast training characteristic of BLS and the advantage of the Jacobi method for fitting high frequencies, which makes it achieve best performance among the three algorithms.

The above numerical simulation results confirm the effectiveness of the proposed BLS-Jacobi method designed according to the frequency principle presenting in BLS.

\section{Conclusion}
Random neural networks, which avoid suffering from  a time-consuming training process, offer an alternative scheme to DNNs. In this paper, we shed some lights into the RFLNNs from a frequency domain perspective, and observe that the frequency principle presenting in the ELM, BLS and stacked BLS: they capture low-frequency components quickly and then gradually fit the high-frequencies. The results further show that for the stacked BLS, the fitting accuracy is not obviously improved when the number of added BLS blocks is more than 2. This may be a defect of the random neural networks with deep structure and is also an interest research topic in the future.

The frequency principle in RFLNNs is of great important to make some improvements and to expand their applications. Based on frequency principle, in this paper we design a method that can generate a BLS network with better prediction performance than the original method, meanwhile, we combine the BLS with the Jacobi iterative method to obtain a more efficient method (BLS-Jacobi) for solving Poisson's equation. The discovered principle of RFLNNs will play an important enlightening role for researchers to develop more potential applications.

\bibliographystyle{IEEEtran}
\bibliography{IEEEexample,bibfile}

\end{document}